\documentclass[conference]{IEEEtran}
\IEEEoverridecommandlockouts
\usepackage{cite}
\usepackage{amsmath,amssymb,amsfonts}
\usepackage{algorithmic}
\usepackage{graphicx}
\usepackage{textcomp}
\usepackage{xcolor}

\usepackage{subfigure}
\usepackage{multirow}
\usepackage{multicol}
\usepackage{booktabs}
\usepackage{url}

\def\BibTeX{{\rm B\kern-.05em{\sc i\kern-.025em b}\kern-.08em
    T\kern-.1667em\lower.7ex\hbox{E}\kern-.125emX}}
\begin{document}

\title{A New Method to Capturing Compositional Knowledge in Linguistic Space}

\author{\IEEEauthorblockN{Jiahe Wan}
\IEEEauthorblockA{\textit{School of Computer Science} \\
\textit{South-Central Minzu University}\\
Wuhan, China \\
2023110239@scuec.edu.cn}
}

\maketitle

\begin{abstract}
Compositional understanding allows visual language models to interpret complex relationships between objects, attributes, and relations in images and text. However, most existing methods often rely on hard negative examples and fine-tuning, which can overestimate improvements and are limited by the difficulty of obtaining hard negatives. In this work, we introduce Zero-Shot Compositional Understanding (ZS-CU), a novel task that enhances compositional understanding without requiring hard negative training data. We propose YUKINO (Yielded Compositional Understanding Knowledge via Textual Inversion with NO), which uses textual inversion to map unlabeled images to pseudo-tokens in a pre-trained CLIP model. We propose introducing ``no'' logical regularization to address the issue of token interaction in inversion. Additionally, we suggest using knowledge distillation to reduce the time complexity of textual inversion. Experimental results show that YUKINO outperforms the existing multi-modal SOTA models by over 8$\%$ on the SugarCREPE benchmark, and also achieves significant improvements in image retrieval tasks.
\end{abstract}


\section{Introduction}
\begin{figure}[t!]
    \centering
    \includegraphics[width=.9\linewidth]{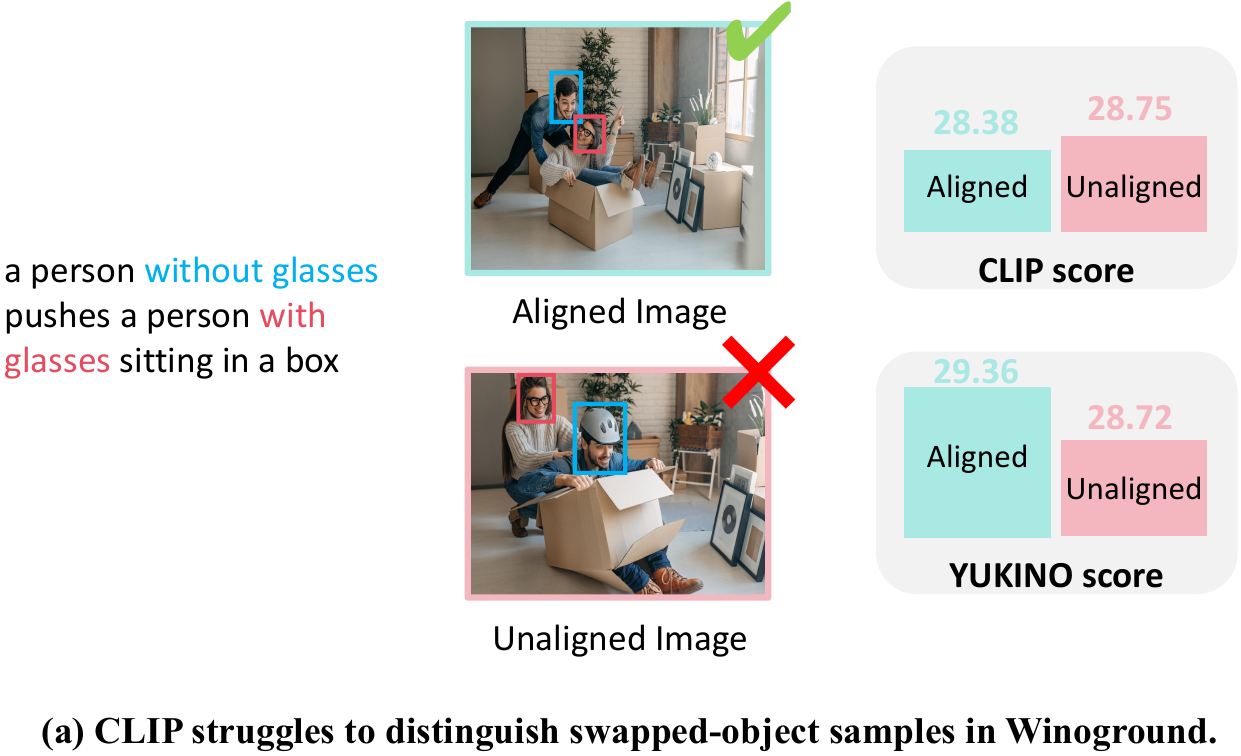} 
    \includegraphics[width=.9\linewidth]{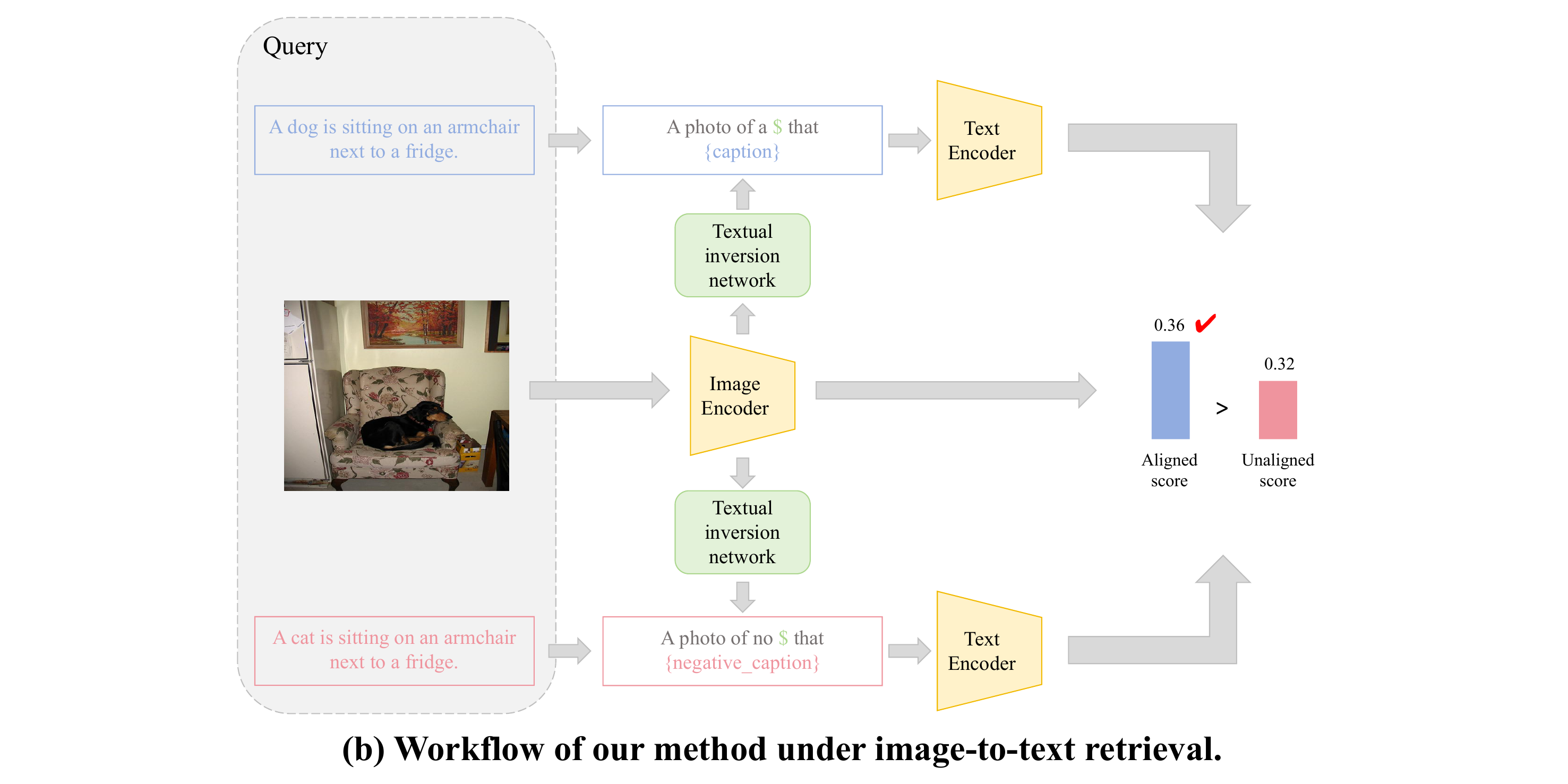}
    \caption{YUKINO introduces a simple enhancement to CLIP that significantly improves its compositional understanding. (a) CLIP and our method YUKINO give similarity scores for captions paired with aligned and unaligned images. (b) Workflow of our proposed method YUKINO under image-to-text retrieval.}
    \label{fig:workflow}
\end{figure}

Compositionality is a fundamental feature common to human vision and natural language. With compositional understanding, people can understand new and more complex scenarios by composing structured representations (i.e., objects, attributes, and relations). For example, compositional understanding allows people to distinguish between ``A person without glasses pushes a person with glasses sitting in a box'' and ``A person with glasses pushes a person without glasses sitting in a box''. However, understanding these scenes remains a difficult task for visual language models. As shown in Fig.~\ref{fig:workflow} (a), the semantic similarity between the caption and the aligned image, as provided by CLIP, is lower than the semantic similarity between the caption and the unaligned image.

Compositional understanding data differ from traditional image-text pairs in that they often contain one or more hard negative samples that are very similar to the original sample. Creating a dataset for compositional understanding is expensive because this type of data is difficult to obtain directly over the Internet. For hard negative text, current studies\cite{crepe2023, aro2023, sugarcrepe2024} can generate hard negative captions corresponding to captions in an automated way, but this may lead to implicit issues such as unreasonable descriptions and lack of fluency\cite{sugarcrepe2024}. However, for hard negative image, current works\cite{winoground2022, cola2023}usually rely on manually curated and small data scale because hard negative images are more difficult to obtain. Current works addressing compositional understanding \cite{tripletclip2024, structureclip2024, negclip2023} rely on supervision to learn the difference between caption and negative caption. For example, Structure-CLIP\cite{structureclip2024} trains CLIP using structural knowledge from scene graphs on the VG-Attribution\cite{aro2023} dataset (a dataset focused on text-based swapped hard negative types), while TripletCLIP\cite{tripletclip2024} enhances compositional reasoning capabilities by introducing a new contrastive pre-training strategy based on generative compositional hard negative samples (containing both text and images). Although current approaches have shown promising results, their reliance on expensively supervised data is such that the models are unintentionally overfitting. 

To eliminate the need for expensive compositional data, we introduce a new task: Zero-Shot Compositional Understanding(ZS-CU). The goal of zero-shot compositional understanding is to design a method that actually improves compositionality without the need for supervised learning, rather than enabling the model to learn to distinguish between the artifacts of hard negatives and positives.

We hypothesize that trained generative models, especially style transfer models, are capable of understanding the complexity of compositional understanding tasks. This hypothesis is based on the idea that generative models can capture high-level visual semantics and pixel-level details, even styles and patterns that are difficult to describe through natural language. To tackle ZS-CU, we propose an approach named Yielded compositional Understanding Knowledge via textual Inversion with NO(YUKINO), which is based on a frozen pre-trained CLIP \cite{clip2021} visual language model. We aim to invert specific images into their corresponding text generation prompts and concatenate them with relevant caption. This text prompt corresponds to a textual representation in the CLIP token embedding space, which we refer to as the pseudo-token. Such pseudo-tokens effectively capture high-level semantics and fine-grained visual representations in the embedding space of the text encoder, allowing objects to be distinguished from other candidate representations through these embeddings. We bootstrap the entire mapping process by text inversion, following the terminology introduced in \cite{animage2023}. YUKINO is obtained through two-phase training on an unlabeled image dataset. In the first phase, our goal is to obtain pseudo-tokens $\$$ that accurately represent the semantic structure of images. We guide the pseudo-tokens to distinguish the structural representation of images by introducing ``no'' logical captions to replace hard negative samples. On the other hand, we apply regularization during the inversion process to address the interaction issues between pseudo-tokens and other text tokens. In the second phase, we distill the above optimization-based textual inversion into a single model $\Theta$, aiming at obtaining a model that can map any image into pseudo-token.

In inference, given a query $(I, T)$, we add ``yes'' prompt and ``no'' prompt to each caption $T$ to generate ``yes'' caption $T^{t}$ and ``no'' caption $T^{n}$. After that we predict the pseudo-token corresponding to $I$ by $\Theta$ and concatenate it to $T^{t}$ and $T^{n}$. To establish a unique match between images and captions, we define the best-matching caption for each image as the one whose ``yes'' version's similarity with the image is greater than the similarity between the ``no'' version of any other caption and the image. Similarly, the best-matching image for each caption is determined similarly. Fig.~\ref{fig:workflow} (b) shows the workflow of the proposed approach. Results on SugarCREPE and Winoground show the state-of-the-art(SOTA) preformance of YUKINO and the effectiveness of its components.

Our contributions can be summarized as follows:
\begin{itemize}
    \item We propose a new task, Zero-Shot Compositional Understanding, to remove the necessity of expensive labeled training data for compositional understanding;
    \item We propose a novel approach called YUKINO that relies on a textual inversion to solve ZS-CU by mapping images to pseudo-tokens. Our approach consists of two phases: optimization-based textual inversion using ``no'' regularization loss and knowledge distillation to obtain a network that can invert everything images;
    \item YUKINO achieves SOTA on SugarCREPE and significantly improves image retrieval capabilities, effectively addressing the issue where CLIP models only distinguish between artifacts of positive and hard negative samples, lacking compositionality.
\end{itemize}

\section{Related Work}
\subsection{Compositional Understanding}
Compositional understanding embodies the ability of a model to match images and texts that have identical word composition.  \cite{aro2023} presents ARO, a benchmark for investigating the sensitivity of VLMs to object order, relations and attributes. This study demonstrates that existing VLMs, despite performing well on downstream tasks, have little compositional understanding,  and presents Neg-CLIP to improve on the investigated shortcomings. Next, CREPE\cite{crepe2023} proposes a benchmark for assessing VLM compositional through systematicity and productivity. SugarCREPE\cite{sugarcrepe2024} refines bias in ARO and CREPE to provide high-quality compositional datasets. To enhance vision-language models' compositional understanding, existing methods suggest training strategies that utilize additional data, models, and/or losses\cite{conclip2024, ilclip2024, tsvlc2023, cocot2024, vq2}. One of the methods that stood out was training the model by augmenting the training data with hard negatives and correct captions\cite{aro2023, structureclip2024}. Although, these methods seem to improve the compositionality of the benchmarks, the models actually utilise the biases in the dataset to achieve this advancement. We analyse this issue in our evaluation.

\subsection{Textual Inversion}
In the field of text-to-image generation, extracting meaningful semantic information from images and mapping it to tokens representing concepts has been proposed as a promising technique for generating highly personalised images\cite{animage2023, multiconcept2023}. Current work\cite{animage2023} proposes a method for performing text inversion using the reconstruction loss of a diffusion model. In addition to the field of text-to-image generation, textual inversion has also been applied to image retrieval tasks\cite{palavra2022, searle2023, isearle2024, pic2word2023}. PALAVRA\cite{palavra2022} pre-trains a mapping function from pre-trained labeled image data and later optimises the concept of input words in the inference phase. Pic2Word\cite{pic2word2023} training textual inversion network on CC3M dataset\cite{cc3m2018} using a cycle contrastive loss. In our work, we capture more fine-grained image semantic information in the text embedding space by introducing "no" logic captions combined with negative regularization.

\begin{figure*}[!htbp]
	\centering
	\includegraphics[width=0.95\linewidth]{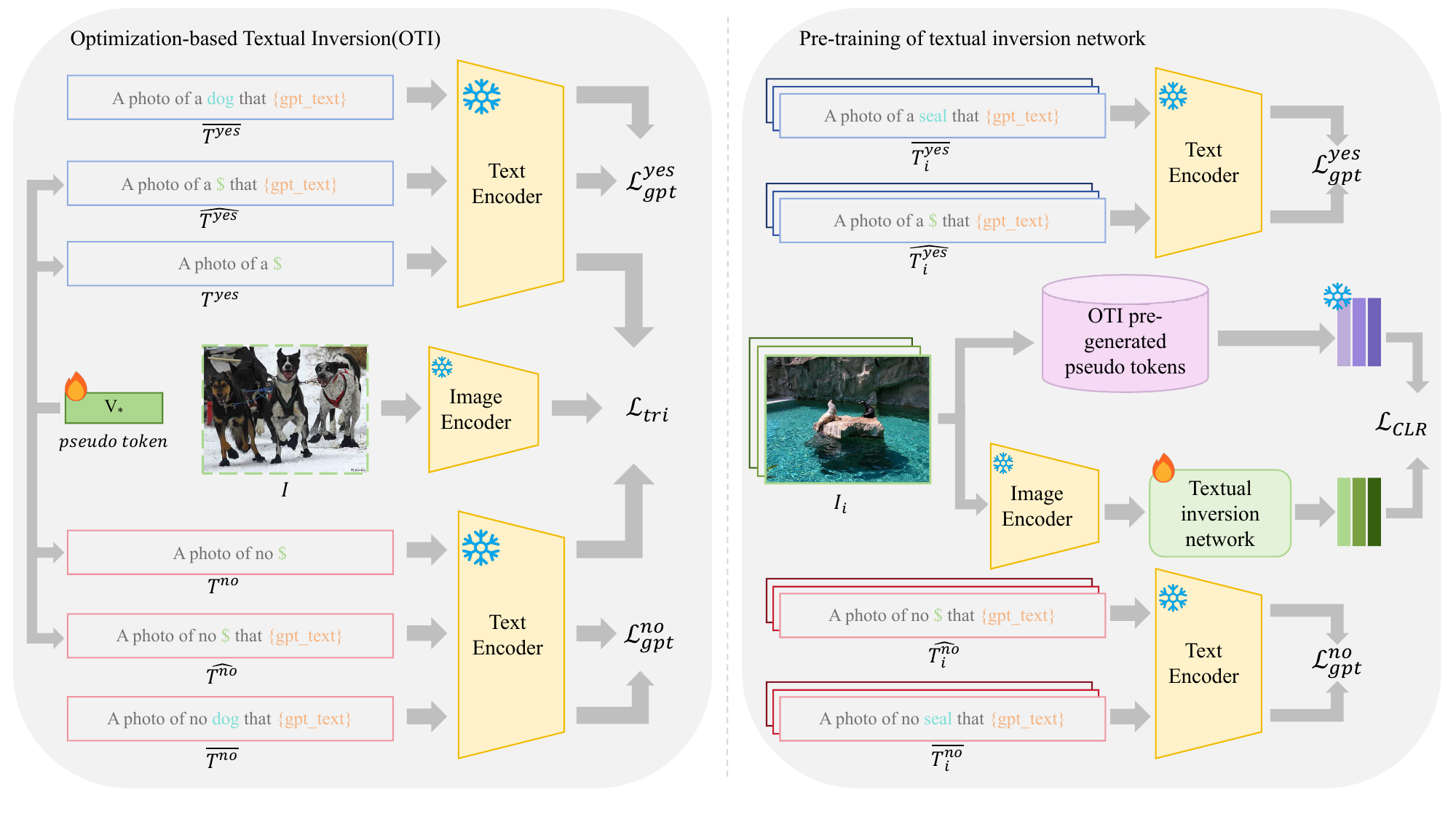}
	\caption{Overview of YUKINO. (a) \emph{Optimization-based Textual Inversion}: we generate a pseudo-token $v_{*}$ from an image \emph{I}(left part).
		(b) \emph{Pre-training of textual inversion network}: we train a network on unlabeled images to gain the ability to quickly invert arbitrary images(right part)
	}
	\label{fig:model}
	\vspace{-10pt}
\end{figure*}

\subsection{Knowledge Distillation}
Knowledge distillation is a technique for transferring knowledge from large, computationally expensive teacher models to smaller student models without reducing validity\cite{distilling2015, distilling2023}. The approach has been successfully applied to several computer vision tasks through self-supervised learning. DINOv2\cite{dinov22024} distils a series of models that outperform OpenCLIP on the basis of a pre-trained ViT model with a billion parameters. \cite{clipkd2024} examines the effectiveness of knowledge distillation for CLIP in terms of relation, feature, gradient and contrastive paradigm. In our work, we extract the knowledge obtained by computationally expensive optimisation-based textual inversion methods into a small neural network via knowledge distillation. In other words, our textual inversion network is an alternative model to resource-intensive optimisation methods.


\section{Methodology}
\subsection{Overview}
Our approach leverages CLIP \cite{clip2021}, a vision-language model trained on a large-scale dataset to align images with their corresponding text within a shared embedding space. CLIP consists of a visual encoder $\phi$ and a text encoder $\psi$. Given an image $\mathbf{x}$, visual features $f = \phi (\mathbf{x})$ by the visual encoder $\phi$. Given a caption $\mathbf{t}$, word embedding layer $\Gamma$ will map each tokenized word into token embedding space, and then the text encoder $\psi$ gets text feature $g = \psi ( \Gamma (\mathbf{t}))$. Our approach aims to generate an image representation that can be used as input to the CLIP text encoder, which we call pseudo-token since it does not correspond to the actual word. These pseudo-tokens effectively capture high-level semantics and fine-grained visual representations within the embedding space of the text encoder, enabling the correct objects to be distinguished from candidate representations through these embeddings.

Inspired by text-conditioned controls in image editing and style transfer, we hypothesize that powerful generative methods can effectively capture fine-grained visual representations. Our approach, named ``YUKINO'', leverages a two-stage training process on unlabeled image datasets to transfer this capability to compositional understanding in a discriminative task setting. First, we obtain pre-generated pseudo-tokens by inverting a set of specific images through an optimization-based textual inversion method (OTI). Second, we train $\Theta$ by knowledge distillation in the pre-generated pseudo-tokens. The text inversion network $\Theta$ takes as input the visual features passing through the CLIP visual encoder and outputs the corresponding pseudo-token. We only use a single pseudo-token to represent the image because \cite{animage2023} shows that a single pseudo-token is sufficient to encode the image information.

Essentially, OTI performs the same operation as $\Theta$, capturing fine-grained visual representations with correct structural information within the shared embedding space. However, due to the iterative nature of OTI, directly using it during inference incurs a non-negligible time overhead. Considering that the effectiveness of OTI in generating pseudo-tokens has been demonstrated (in Section \ref{sec:experiments}), we propose to distill the knowledge of OTI into a feed-forward network that has both the strong expressive power of OTI and a negligible time overhead. We refer to our approach as YUKINO, and when we use only OTI, we call our approach YUKINO-OTI.

\subsection{Optimization-based Textual Inversion (OTI)}
Given an image $I$, we find $v_{*}$ through direct optimization. Specifically, we represent the structural information to be captured from the image as a character $\$$. We intervene in the embedding process, and replace the vector associated with the labeled character with the newly learned embedding $v_{*}$. Fundamentally, it is the infusion of visual information into our vocabulary, where we can form new sentences that contain image information, just as we would with any other word. Fig.~\ref{fig:model} provides an overview of OTI.

To regulate the optimization process, we randomly sample the context $T^{yes}$ originating from the PALAVRA\cite{palavra2022} template, e.g., ``A photo of a \$'', and feed it into the CLIP text encoder $\psi$ to obtain $g^{yes} = \psi(T^{yes})$. In order to equip pseudo-tokens with the ability to distinguish between structured representations rather than simply mapping global visual representations, we introduce ``no'' logic captions instead of using hard negative captions (such as those involving object swaps within captions), e.g., ``A photo of no \$'', which is fed into the CLIP text encoder $\psi$, to obtain the negative text feature $g^{no} = \psi(T ^{no})$. Given an image $I$, we obtain image features $f = \phi(I)$ from the CLIP image encoder.

Our goal is to obtain a pseudo-token $v_{*}$ that has compositionality about the image $I$, so we make the distance between the image and the positive textual features smaller than the distance between the image and the negative text in the common embedding space of CLIP. To achieve our aim, we utilize CLIP-based triplet loss:
\begin{equation}
    \mathcal{L}_{tri} = max\{ \left \| f - g^{yes} \right \| _{2} - \left \| f - g^{no} \right \| _{2} + 1, 0\}
\end{equation}

However, since image features and word token embeddings are located in their own spaces, $\mathcal{L}_{tri}$ forces the pseudo-token into sparse regions of the CLIP token embedding space. To ensure that the pseudo-token does not collapse into a meaningless representation, we use regularization techniques to constrain it within the CLIP token embedding manifold, enhancing its interaction with other text token and thereby improving its reasoning ability.

For an unlabeled image $I$, we identify multiple objects most similar to it and randomly sample one during each iteration. A GPT-prompted caption $\overline{T^{yes}}$ is created using the sampled object, while replacing the object with $\$$ generates the pseudo-caption $\widehat{T^{yes}}$. CLIP's text encoder is then used to extract features $\overline{g^{yes}} = \psi(\overline{T^{yes}})$ and $\widehat{g^{yes}} = \psi(\widehat{T^{yes}})$, aligning them by minimizing their cosine loss:
\begin{equation}
    \mathcal{L}^{yes}_{gpt} = 1 - \cos \left(\overline{g^{yes}},\widehat{g^{yes}} \right)
\end{equation}

For ``no'' logic texts, we minimize the distance between $\overline{g^{no}}$ and $\widehat{g^{no}}$ using the same loss. This loss takes into account the context of the text while $v^{*}$ associates the semantics of the image, enhancing the ability of $v^{*}$ to interact with other text tokens. During the OTI, our total losses were:
\begin{equation} \label{eq.otiloss}
    \mathcal{L}_{OTI} = \lambda_{tri} \mathcal{L}_{tri} + \lambda_{OTIgpt} \left(\mathcal{L}^{yes}_{gpt} + \mathcal{L}^{no}_{gpt} \right) 
\end{equation}
where $\lambda_{tri}$ and $\lambda_{OTIgpt}$ are the loss weights.

\begin{table*}[!htbp]
    \centering
    \small
    \caption{Results ~($\%$) on Vision-language compositionality benchmarks as image-to-text retrieval problem. The Benchmark is SugarCREPE. Best and second-best scores are highlighted in bold and underlined, respectively. $\dag$ indicates results from the original paper, and the results of other methods are reproduced by us.}
    \label{tab:sugarcrepe}
\begin{tabular}{lc|ccccccc|ccc}
    \toprule
    \multirow{2}{*}{Dataset} & \multirow{2}{*}{Method} & \multicolumn{3}{c}{REPLACE} & \multicolumn{2}{c}{SWAP} & \multicolumn{2}{c|}{ADD} & \multicolumn{3}{c}{Avg}\\
    \cline{3-12}
    & & obj & att & rel & obj & att & obj & att & REPLACE & SWAP & ADD \\
    \midrule
    \multirow{5}{*}{LAION2B} 
    & CLIP\cite{clip2021} & 93.77 & 82.49 & 68.92 &  60.00 & 67.42 & 87.10 & 77.89 & 81.73 & 63.71 & 82.50 \\
    & Neg-CLIP\cite{negclip2023} & 92.62 & 84.64 & 71.41 & 74.69 & 76.58 & 86.71 & 85.26 & 82.89 & 75.64 & 85.99 \\
    & Structure-CLIP\cite{structureclip2024} & 91.77 & 86.8 & 75.32 & 74.29 & 84.83 & 88.94 & 88.01 & 84.63 & 79.56 & 88.48 \\
    & YUKINO-OTI (Ours) & \textbf{97.82} & \textbf{95.43} & \textbf{91.68} & \underline{87.76} & \textbf{90.69} & \underline{96.51} & \textbf{94.80} & \textbf{94.98} & \textbf{89.23} & \textbf{95.66} \\
    & YUKINO (Ours) & \underline{97.15} & \underline{94.80} & \underline{91.25} & \textbf{88.57} & \underline{89.79} & \textbf{96.65} & \underline{93.21} & \underline{94.40}  & \underline{89.18}  & \underline{94.93}  \\
    \midrule
    \multirow{2}{*}{CC12M} 
    & Neg-CLIP$^\dagger$ \cite{negclip2023} & 77.84 & 69.29 & 63.23 & 66.53 & 62.31 & 67.71 & 69.65 & 70.12 & 64.42 & 68.41 \\
    & TripletCLIP$^\dagger$ \cite{tripletclip2024} & 83.66 & 81.22 & 79.02 & 64.49 & 63.66 & 73.67 & 75.43 & 81.30 & 64.08 & 74.55 \\
    \bottomrule
\end{tabular}
\end{table*}

\subsection{Pre-training of textual inversion network}

The pseudo-tokens extracted by OTI are highly effective, capturing the image's structural information and functioning like words for easy application. To address the long execution time caused by OTI inverting specific images, we designed a surrogate model through knowledge distillation, which can invert any image. In this work, we adopt a simple network with three linear layers, referred to as $\Theta$.

During training, we define a small batch with $N$ input pairs as $\mathcal{B} = \{ I^{i} \}^{N}_{i=1} \in \mathbb{D}$, where $\mathbb{D}$ is an unlabeled image datasets. On the one hand, we use the OTI to map each image to a pseudo-token, which ultimately leads to a set of pseudo-tokens $V_{*} = \{ v^{i}_{*} \}^{N}_{i=1}$. Although this process is time-consuming, it is a one-time event and we find it acceptable. On the other hand, we map the visual features of the image to the predicted pseudo-token through $\Theta$, i.e., $\tilde{v}^{i}_{*} = \Theta(f^{i})$. We optimize this process using symmetric contrastive loss as follows:
\begin{equation}
    \mathcal{L}_{CLR} = - \frac{1}{N} \sum^{N}_{k=1} \left [ \ell \left(\tilde{v}^{k}_{*},v^{k}_{*},v^{j}_{*} \right) + \ell \left(v^{k}_{*},\tilde{v}^{k}_{*},\tilde{v}^{j}_{*} \right) \right ]
\end{equation}
where $\ell$ is used to maximize the cosine distance between the pre-generated pseudo-token $v^{i}_{*}$ and the $\Theta$ predicted pseudo-token $\tilde{v}^{i}_{*}$ while minimizing the similarity between different tokens, formulated as follow:
\begin{equation} \label{eq:loss}
    \ell \left(\alpha,\beta,\gamma \right) = \log \frac{\exp \left(\frac{\cos \left(\alpha,\beta \right)}{\tau} \right)}{\sum^{N}_{j=1} \exp \left(\frac{\cos \left(\beta,\gamma \right)}{\tau} \right) + \sum_{j \ne k} \exp \left(\frac{\cos \left(\alpha,\gamma \right)}{\tau} \right)}
\end{equation}
where $\tau$ is a temperature parameter.

To normalize the training of $\Theta$, we follow the same regularization technique described in OTI. The total loss function used to update the weights of $\Theta$ is:
\begin{equation} \label{eq:YUKINOloss}
    \mathcal{L}_{YUKINO} = \lambda_{CLR} \mathcal{L}_{CLR} + \lambda_{gpt} \left(\mathcal{L}^{yes}_{gpt} + \mathcal{L}^{no}_{gpt} \right)
\end{equation}
where $\lambda_{CLR}$ and $\lambda_{gpt}$ are the loss weights.


\section{Experiments}
\label{sec:experiments}
\subsection{Datasets and Evaluation Metrics}
\subsubsection{Pretraining Datasets}
We trained CLIP on the same pretraining dataset, LAION-2B\cite{laion2022}, and used it as the backbone to reproduce the baselines following the methods in the original papers. For the recently released TripletCLIP\cite{tripletclip2024}, since the training code has not been made publicly available, we directly used the original data from their paper and compared it with NegCLIP\cite{negclip2023}, trained on the same pretraining dataset.

\subsubsection{Downstream Datasets}
The main goal of this study is to enhance CLIP's compositional capability; therefore, OpenCLIP-ViT-B/32 is used as the backbone model in the experiments. We evaluate YUKINO and baseline models using the SugarCREPE and Winoground compositional benchmarks. Additional backbone experiment results are provided in the appendix. 

\subsubsection{Evaluation Metrics}
 The evaluation metrics are based on accuracy, but the Winoground benchmark demands stronger compositional understanding. It requires the model to accurately distinguish the correct pairings in the mixed matching combinations, where the Group Score demands both text retrieval and image retrieval to be correct simultaneously. The evaluation follows the methods outlined in CLIP-Benchmark\footnote{\url{https://github.com/LAION-AI/CLIP_benchmark}} or the official benchmarks.

\subsection{Implementation Details}
All of our experiments are performed on $4\times$ NVIDIA A6000 GPU with the Pytorch framework. We use the unlabeled test set of ImageNet1K\cite{imagenet1k} as the unlabeled dataset to optimize OTI and train $\Theta$. The loss weights $\lambda_{tri}$ and $\lambda_{OTIgpt}$ in \eqref{eq.otiloss} are set to 1 and 0.5, respectively. The loss weights $\lambda_{CLR}$ and $\lambda_{gpt}$ in \eqref{eq:YUKINOloss} are set to 1 and 0.5. We use GPT-Neo-2.7B\cite{gpt2020} to generate ``\{gpt\_text\}''. Due to space constraints, we provide the implementation details in the supplementary material.

\begin{figure*}[htbp]
    \centering
    \subfigure[CLIP]{
        \includegraphics[width=0.22\textwidth]{"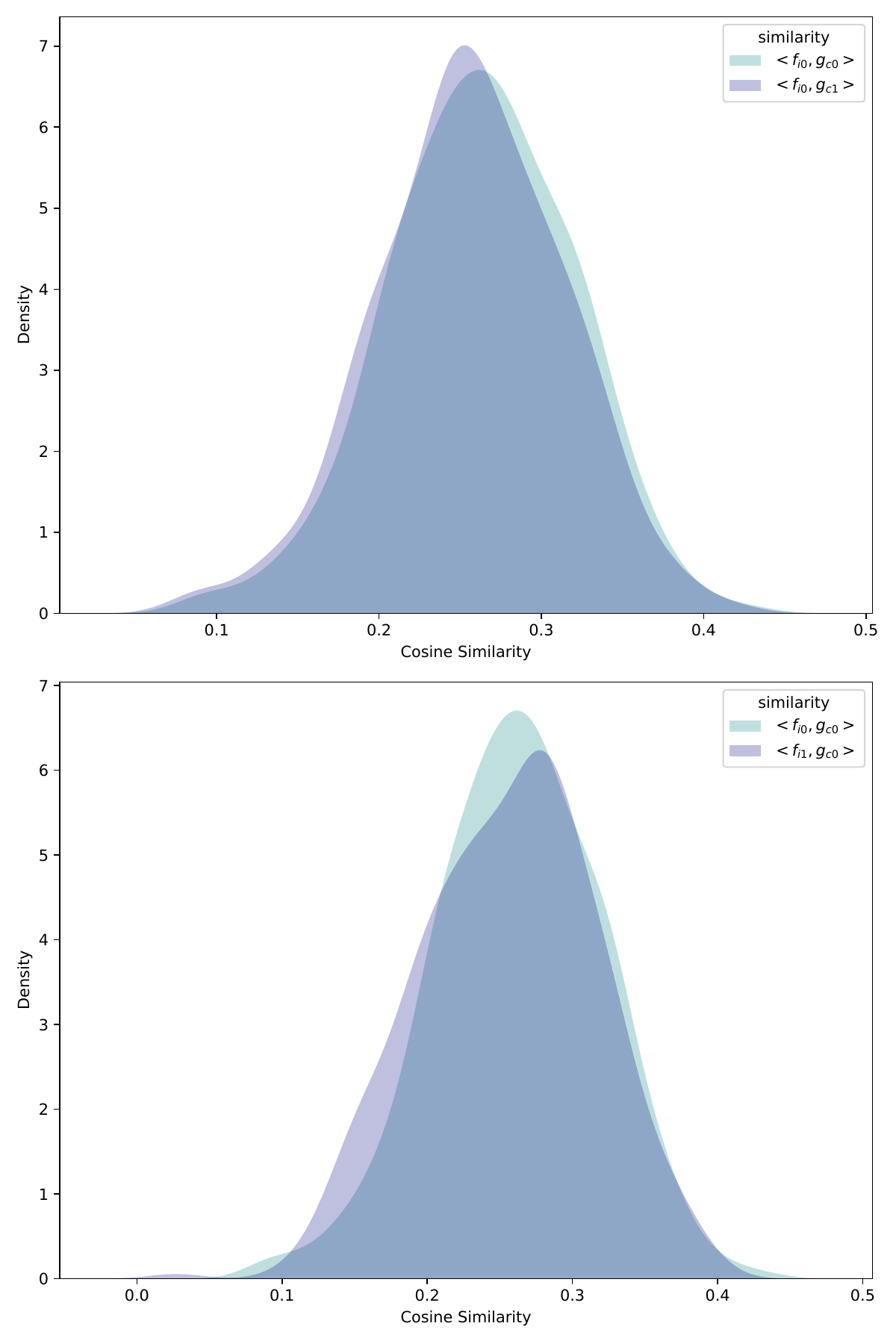"}
    }
    \subfigure[Neg-CLIP]{
        \includegraphics[width=0.22\textwidth]{"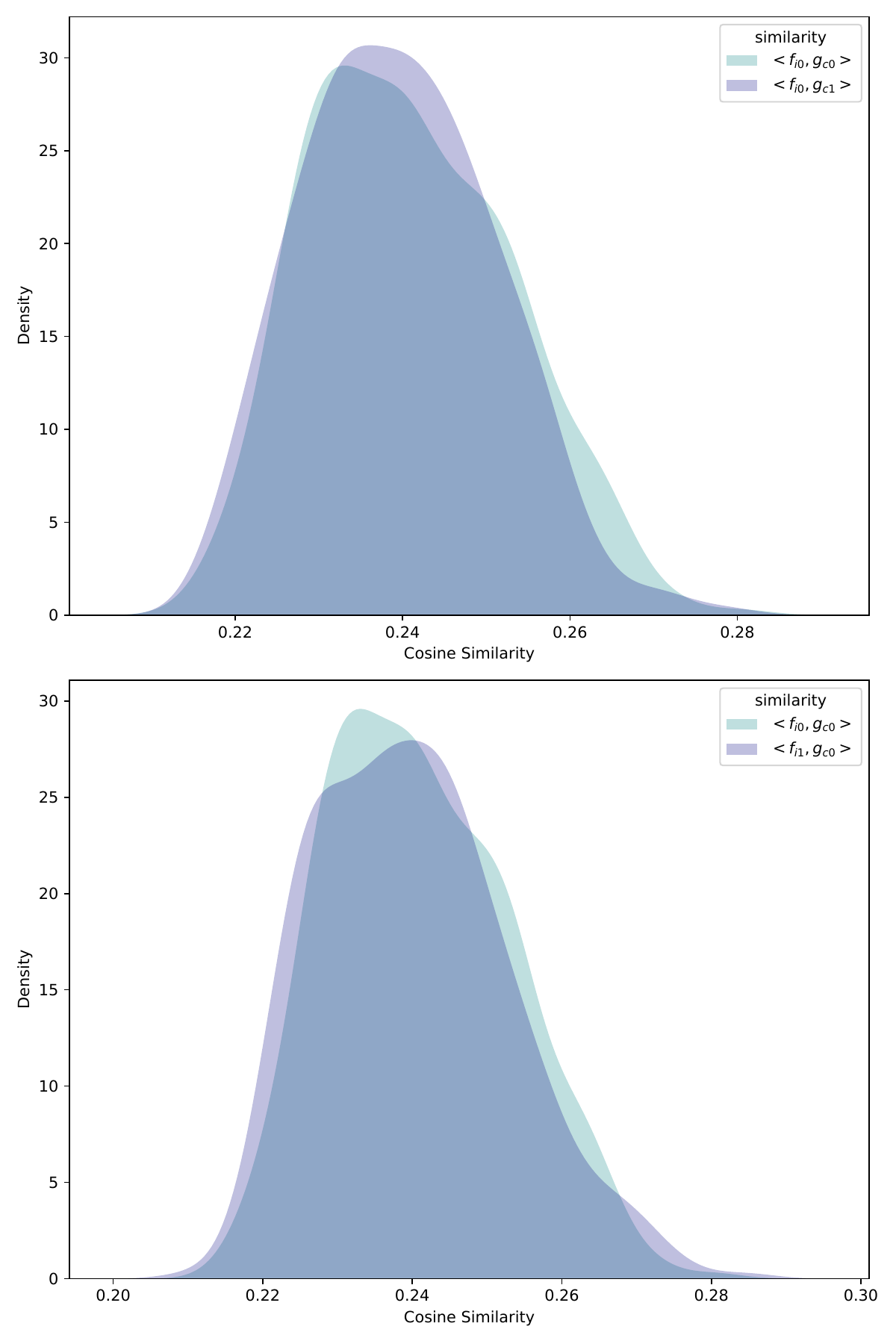"}
    }
    \subfigure[Structure-CLIP]{
        \includegraphics[width=0.22\textwidth]{"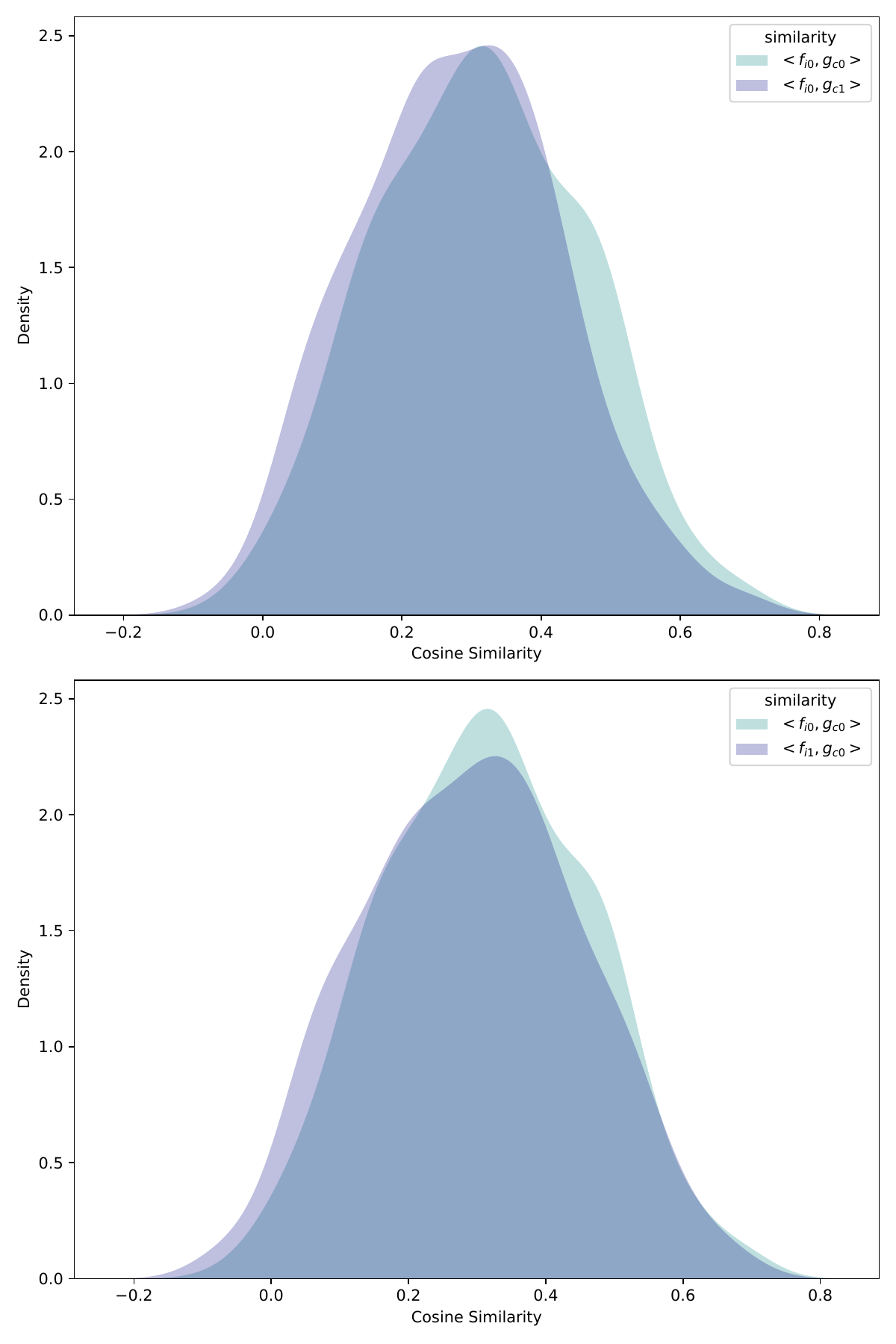"}
    }
    \subfigure[YUKINO]{
        \includegraphics[width=0.22\textwidth]{"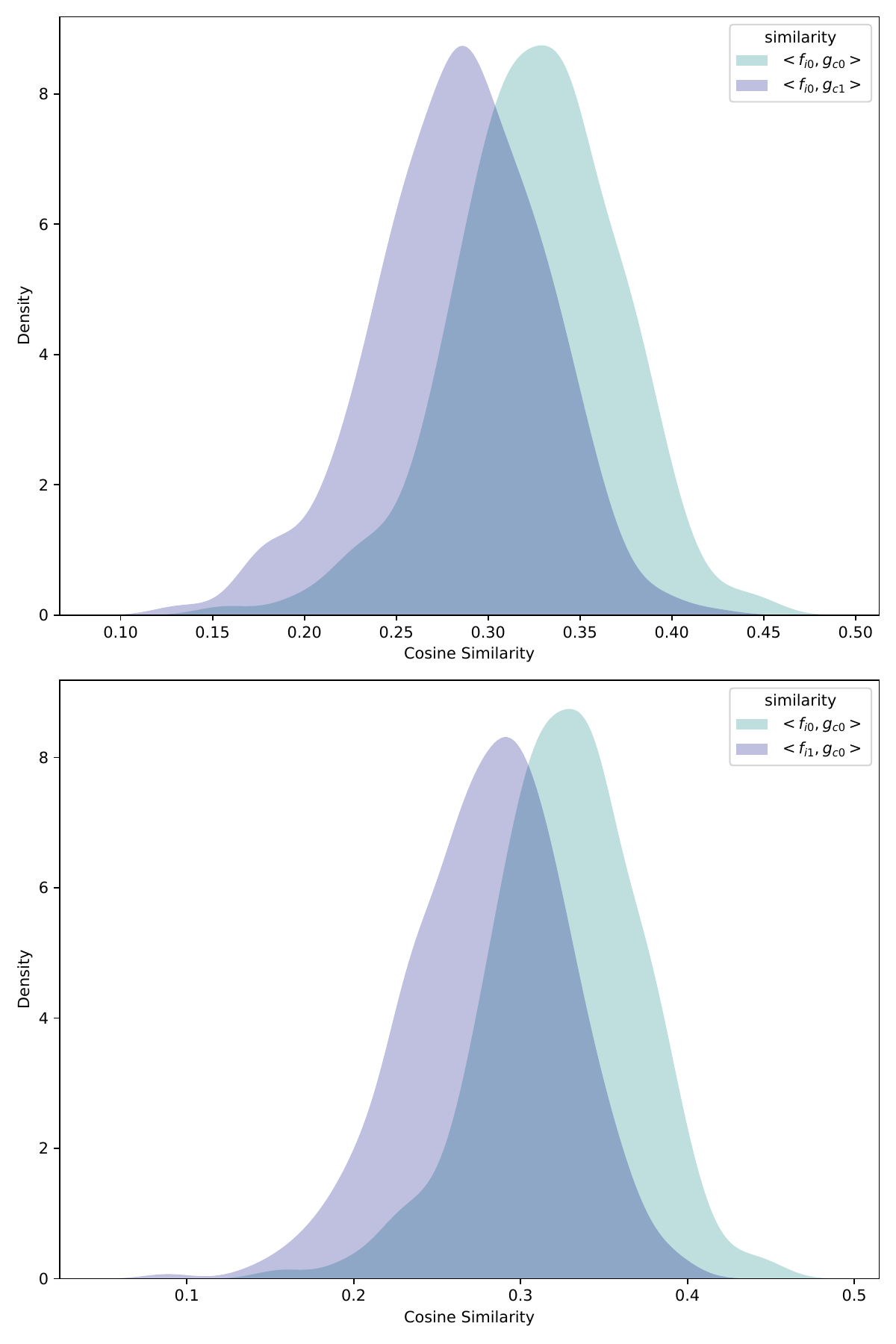"}
    }
    \caption{Similarity density of CLIP, Neg-CLIP, Structure-CLIP and YUKINO. (a), (b), (c) and (d) show the similarity density of image0 with all captions and caption0 with all images in Winoground for the 4 models, respectively.}
    \label{fig:density}
\end{figure*}

\subsection{Comparison Results with State-of-the-art Methods}
\subsubsection{Quantitative comparison on SugarCREPE}
We present the quantitative results of our method compared with recent state-of-the-art methods on the SugarCREPE benchmark, as shown in Table~\ref{tab:sugarcrepe}. Our method consistently outperforms all models based on the same CLIP framework. On average, YUKINO using ViT-B-32 outperforms Structure-CLIP by 9.77$\%$ on REPLACE, 9.62$\%$ on SWAP and 6.45$\%$ on ADD, respectively. In the REPLACE and ADD categories, TripletCLIP outperforms NegCLIP by an average of 11.18$\%$ and 6.14$\%$, respectively. In contrast, our YUKINO surpasses NegCLIP by an average of 11.51$\%$, 13.54$\%$ and 8.94$\%$ across the three aspects, further highlighting the superiority of our method. It is worth noting that the iterative OTI inversion of an image takes an average of 108 seconds, while the text inversion network only takes 2 seconds to invert an image. With a significant reduction in text inversion time, YUKINO demonstrates stronger compositional ability than YUKINO-OTI in the more challenging swap-obj task, while performing similarly to YUKINO-OTI in other aspects. 

\begin{table}[htbp]
    \centering
    \small
    \caption{comparison of results ~($\%$) on Winoground benchmark }
    \label{tab:winoground}
\begin{tabular}{lc|ccc}
    \toprule
    Dtaset & Method & Text & Image & Group \\
    \midrule
    \multirow{5}{*}{LAION2B}
    & CLIP\cite{clip2021} & 34.75 & 11.00 & 7.50 \\
    & Neg-CLIP\cite{negclip2023} & 20.50 & 12.25 & 6.50 \\
    & Structure-CLIP\cite{structureclip2024} & 22.50 & 14.00 & 8.25 \\
    & YUKINO-OTI (Ours) & \underline{66.75} & \underline{42.50} & \underline{37.50} \\
    & YUKINO (Ours) & \textbf{74.00} & \textbf{46.25} & \textbf{42.75} \\
    \midrule
    \multirow{2}{*}{CC12M}
    & Neg-CLIP$^\dagger$\cite{negclip2023} & 18.25 & 6.50 & 4.25 \\
    & TripletCLIP$^\dagger$\cite{tripletclip2024} & 23.25 & 6.25 & 4.25 \\
    \midrule
    \multirow{3}{*}{-}
    & LLaVA-1.5\cite{llava1.5} & 36.00 & 33.25 & 20.05 \\
    & GPT-4V\cite{cocot2024} & 54.50 & 42.50 & 37.75 \\
    & GPT-4V+CoCoT$^\dagger$\cite{cocot2024} & \textbf{58.50} & \textbf{49.50} & \textbf{44.50} \\
    \bottomrule
\end{tabular}
\end{table}

\subsubsection{Quantitative comparison on Winoground}
In Table~\ref{tab:winoground}, we report the results on the Winoground benchmark. Compared to SugarCREPE, our method demonstrates more substantial improvements on Winoground, excelling in image retrieval. The reason why YUKINO outperforms YUKINO-OTI is that YUKINO learns from pseudo-tokens pre-generated on a larger dataset, which enhances its ability to capture visual semantics. Table~\ref{tab:winoground} also compares YUKINO with various multimodal large language models, which involve more parameters and training data, showing that YUKINO achieves comparable performance to GPT-4V with chain-of-thought reasoning (GPT-4V+CoCoT) in image retrieval.

\subsection{Representation Distribution Analysis}
We observe that models fine-tuned with hard negatives exhibit significantly different performance on SugarCREPE and Winoground. On the SugarCREPE benchmark, the model only needs to learn to identify the caption that better matches the image between the hard negative caption and the positive caption to improve its score. However, on Winoground, the model must correctly match each image-text pair within a set of mixed image-text combinations.

We attribute this stark difference to the fact that models like Neg-CLIP, fine-tuned with hard negatives, only learn to distinguish the artifacts between hard negative instances and positive instances without improving compositionality. We conducted an experiment to verify our conclusions. Specifically, we extracted image and text features from Winoground and calculated their cosine similarities. Finally, we used kernel density estimation to estimate the density of these similarities, as shown in Fig.~\ref{fig:density}. The ideal outcome should be that image0 features matches caption0 features. From Fig.~\ref{fig:density}, we find that the original CLIP, Neg-CLIP and Structure-CLIP fails to achieve it because all similarities are mixed together. On the contrary, YUKINO has the smallest overlapping region for both similarity densities and has higher values on $<f_{i0},g_{c0}>$ and lower values on $<f_{i0},g_{c1}>$ and $<f_{i1},g_{c0}>$. This suggests that YUKINO can enhance the compositionality of vision-language models by capturing visual details that are difficult to express in natural language.

\begin{table}[htbp]
    \centering
    \caption{Ablation results on Winoground. }
    \label{tab:abstudies}
\begin{tabular}{cc|ccc}
    \toprule
    \multicolumn{1}{c}{Abl.} & \multicolumn{1}{c}{Method} & \multicolumn{1}{|c}{Text} & \multicolumn{1}{c}{Image} & \multicolumn{1}{c}{Group} \\
    \midrule
    \multirow{4}{*}{OTI} & w/o reg & 51.25 & 30.50 & 24.00 \\
    & w/o ``yes'' reg & 56.75 & 35.50 & 31.25 \\
    & w/o ``no'' reg & \textbf{71.75} & \textbf{44.00} & \underline{36.75} \\
    & YUKINO-OTI & \underline{66.75} & \underline{42.50} & \textbf{37.50} \\
    \midrule
    \multirow{5}{*}{$\Theta$} & cos distill & 52.75 & 26.50 & 23.00 \\
    & w/o reg & 64.25 & 42.00 & 37.75 \\
    & w/o ``yes'' reg & \underline{66.00} & \underline{42.25} & \underline{38.25} \\
    & w/o ``no'' reg & \underline{66.00} & 36.50 & 33.25 \\
    & YUKINO & \textbf{74.00} & \textbf{46.25} & \textbf{42.75} \\
    \midrule
    LM & YUKINO-llama & \textbf{74.00} & 45.50 & 42.25 \\
    \midrule
    Datasets & YUKINO-VG & 73.00 & \textbf{48.00} & \textbf{45.00} \\
    \bottomrule
\end{tabular}  
\end{table}

\subsection{Ablation Studies}
\subsubsection{Analysis of the impact of different language model}
We evaluate the effectiveness of different language models in capturing visual information by replacing GPT-Neo-2.7B with LLama3-8B \cite{llama32024} to generate ``{gpt\_text}''. The version using LLama3's $\Theta$ variant is referred to as YUKINO-llama. YUKINO-llama performs similarly to YUKINO in Text Score but slightly worse in Image Score and Group Score. While LLama3-8B generates more detailed captions, it still relies on image features during training, and YUKINO-llama does not add additional visual context, limiting its ability to capture richer visual semantics.

\subsubsection{Analysis of the impact of different datasets}
We also trained a variant of $\Theta$ using the Visual Genome\cite{visualgenome} training dataset, referred to as YUKINO-VG. Visual Genome offers more structured image information than ImageNet1K, providing a richer context for the model. Importantly, we use only the raw images, keeping the approach unsupervised. Our method further enhances CLIP's compositional ability in this richer visual context, consistent with previous analysis.

\subsubsection{Analysis of the impact of regularization loss}
We demonstrate the effectiveness of the regularization loss through experiments. \textit{w/o reg}, \textit{w/o ``yes'' reg}, and \textit{w/o ``no'' reg} represent no regularization loss, using only the ``no'' regularization loss, and using only the ``yes'' regularization loss, respectively. We can see that both regularization losses can improve the role of pseudo-token in compositional understanding. Although using only the ``yes'' regularization achieves higher performance in a single aspect, we believe that overall performance is more important in more complex scenarios, making it necessary to use both regularizations simultaneously.

\subsubsection{Analysis of the impact of distillation loss}
Compared to the cosine version of the distillation loss (\textit{cos distill}), the contrastive version of the distillation loss significantly improves performance. This suggests that learning from OTI pre-generated tokens is more effective than learning from the original image. 

\section{Conclusions}
In this paper, we propose a method to address zero-shot compositional understanding. Our model aims to (1) eliminate the dependency on expensive compositional data and (2) capture high-level visual semantics from images to enhance the model's compositional understanding ability. Our approach embeds complex visual representations into natural language text for true compositional understanding, rather than relying on artificial patterns between samples.


\bibliographystyle{IEEEbib}
\bibliography{icme2025references}

\begin{thebibliography}{10}

\bibitem{crepe2023}
Zixian Ma, Jerry Hong, Mustafa~Omer Gul, Mona Gandhi, Irena Gao, and Ranjay
  Krishna,
\newblock ``Crepe: Can vision-language foundation models reason
  compositionally?,''
\newblock in {\em Proceedings of the IEEE/CVF Conference on Computer Vision and
  Pattern Recognition}, 2023, pp. 10910--10921.

\bibitem{aro2023}
Mert Yuksekgonul, Federico Bianchi, Pratyusha Kalluri, Dan Jurafsky, and James
  Zou,
\newblock ``When and why vision-language models behave like bags-of-words, and
  what to do about it?,''
\newblock in {\em The Eleventh International Conference on Learning
  Representations}, 2022.

\bibitem{sugarcrepe2024}
Cheng-Yu Hsieh, Jieyu Zhang, Zixian Ma, Aniruddha Kembhavi, and Ranjay Krishna,
\newblock ``Sugarcrepe: Fixing hackable benchmarks for vision-language
  compositionality,''
\newblock {\em Advances in Neural Information Processing Systems}, vol. 36,
  2024.

\bibitem{winoground2022}
Tristan Thrush, Ryan Jiang, Max Bartolo, Amanpreet Singh, Adina Williams, Douwe
  Kiela, and Candace Ross,
\newblock ``Winoground: Probing vision and language models for visio-linguistic
  compositionality,''
\newblock in {\em Proceedings of the IEEE/CVF Conference on Computer Vision and
  Pattern Recognition}, 2022, pp. 5238--5248.

\bibitem{cola2023}
Arijit Ray, Filip Radenovic, Abhimanyu Dubey, Bryan~A Plummer, Ranjay Krishna,
  and Kate Saenko,
\newblock ``Cola: How to adapt vision-language models to compose objects
  localized with attributes?,''
\newblock {\em arXiv preprint arXiv:2305.03689}, 2023.

\bibitem{tripletclip2024}
Maitreya Patel, Abhiram Kusumba, Sheng Cheng, Changhoon Kim, Tejas Gokhale,
  Chitta Baral, and Yezhou Yang,
\newblock ``Tripletclip: Improving compositional reasoning of clip via
  synthetic vision-language negatives,''
\newblock {\em arXiv preprint arXiv:2411.02545}, 2024.

\bibitem{structureclip2024}
Yufeng Huang, Jiji Tang, Zhuo Chen, Rongsheng Zhang, Xinfeng Zhang, Weijie
  Chen, Zeng Zhao, Zhou Zhao, Tangjie Lv, Zhipeng Hu, et~al.,
\newblock ``Structure-clip: Towards scene graph knowledge to enhance
  multi-modal structured representations,''
\newblock in {\em Proceedings of the AAAI Conference on Artificial
  Intelligence}, 2024, pp. 2417--2425.

\bibitem{negclip2023}
Roei Herzig, Alon Mendelson, Leonid Karlinsky, Assaf Arbelle, Rogerio Feris,
  Trevor Darrell, and Amir Globerson,
\newblock ``Incorporating structured representations into pretrained vision \&
  language models using scene graphs,''
\newblock {\em arXiv preprint arXiv:2305.06343}, 2023.

\bibitem{clip2021}
Alec Radford, Jong~Wook Kim, Chris Hallacy, Aditya Ramesh, Gabriel Goh,
  Sandhini Agarwal, Girish Sastry, Amanda Askell, Pamela Mishkin, Jack Clark,
  et~al.,
\newblock ``Learning transferable visual models from natural language
  supervision,''
\newblock in {\em International conference on machine learning}. PMLR, 2021,
  pp. 8748--8763.

\bibitem{animage2023}
Rinon Gal, Yuval Alaluf, Yuval Atzmon, Or~Patashnik, Amit~Haim Bermano, Gal
  Chechik, and Daniel Cohen-or,
\newblock ``An image is worth one word: Personalizing text-to-image generation
  using textual inversion,''
\newblock in {\em The Eleventh International Conference on Learning
  Representations}, 2023.

\bibitem{conclip2024}
Jaisidh Singh, Ishaan Shrivastava, Mayank Vatsa, Richa Singh, and Aparna
  Bharati,
\newblock ``Learn" no" to say" yes" better: Improving vision-language models
  via negations,''
\newblock {\em arXiv preprint arXiv:2403.20312}, 2024.

\bibitem{ilclip2024}
Chenhao Zheng, Jieyu Zhang, Aniruddha Kembhavi, and Ranjay Krishna,
\newblock ``Iterated learning improves compositionality in large
  vision-language models,''
\newblock in {\em Proceedings of the IEEE/CVF Conference on Computer Vision and
  Pattern Recognition}, 2024, pp. 13785--13795.

\bibitem{tsvlc2023}
Sivan Doveh, Assaf Arbelle, Sivan Harary, Eli Schwartz, Roei Herzig, Raja
  Giryes, Rogerio Feris, Rameswar Panda, Shimon Ullman, and Leonid Karlinsky,
\newblock ``Teaching structured vision \& language concepts to vision \&
  language models,''
\newblock in {\em Proceedings of the IEEE/CVF Conference on Computer Vision and
  Pattern Recognition}, 2023, pp. 2657--2668.

\bibitem{cocot2024}
Daoan Zhang, Junming Yang, Hanjia Lyu, Zijian Jin, Yuan Yao, Mingkai Chen, and
  Jiebo Luo,
\newblock ``Cocot: Contrastive chain-of-thought prompting for large multimodal
  models with multiple image inputs,''
\newblock {\em arXiv preprint arXiv:2401.02582}, 2024.

\bibitem{vq2}
Michal Yarom, Yonatan Bitton, Soravit Changpinyo, Roee Aharoni, Jonathan
  Herzig, Oran Lang, Eran Ofek, and Idan Szpektor,
\newblock ``What you see is what you read? improving text-image alignment
  evaluation,''
\newblock {\em Advances in Neural Information Processing Systems}, vol. 36,
  2024.

\bibitem{multiconcept2023}
Nupur Kumari, Bingliang Zhang, Richard Zhang, Eli Shechtman, and Jun-Yan Zhu,
\newblock ``Multi-concept customization of text-to-image diffusion,''
\newblock in {\em Proceedings of the IEEE/CVF Conference on Computer Vision and
  Pattern Recognition}, 2023, pp. 1931--1941.

\bibitem{palavra2022}
Niv Cohen, Rinon Gal, Eli~A Meirom, Gal Chechik, and Yuval Atzmon,
\newblock ``“this is my unicorn, fluffy”: Personalizing frozen
  vision-language representations,''
\newblock in {\em European conference on computer vision}. Springer, 2022, pp.
  558--577.

\bibitem{searle2023}
Alberto Baldrati, Lorenzo Agnolucci, Marco Bertini, and Alberto Del~Bimbo,
\newblock ``Zero-shot composed image retrieval with textual inversion,''
\newblock in {\em Proceedings of the IEEE/CVF International Conference on
  Computer Vision}, 2023, pp. 15338--15347.

\bibitem{isearle2024}
Lorenzo Agnolucci, Alberto Baldrati, Marco Bertini, and Alberto Del~Bimbo,
\newblock ``isearle: Improving textual inversion for zero-shot composed image
  retrieval,''
\newblock {\em arXiv preprint arXiv:2405.02951}, 2024.

\bibitem{pic2word2023}
Kuniaki Saito, Kihyuk Sohn, Xiang Zhang, Chun-Liang Li, Chen-Yu Lee, Kate
  Saenko, and Tomas Pfister,
\newblock ``Pic2word: Mapping pictures to words for zero-shot composed image
  retrieval,''
\newblock in {\em Proceedings of the IEEE/CVF Conference on Computer Vision and
  Pattern Recognition}, 2023, pp. 19305--19314.

\bibitem{cc3m2018}
Piyush Sharma, Nan Ding, Sebastian Goodman, and Radu Soricut,
\newblock ``Conceptual captions: A cleaned, hypernymed, image alt-text dataset
  for automatic image captioning,''
\newblock in {\em Proceedings of the 56th Annual Meeting of the Association for
  Computational Linguistics (Volume 1: Long Papers)}, 2018, pp. 2556--2565.

\bibitem{distilling2015}
Geoffrey Hinton, Oriol Vinyals, and Jeff Dean,
\newblock ``Distilling the knowledge in a neural network,''
\newblock {\em arXiv preprint arXiv:1503.02531}, 2015.

\bibitem{distilling2023}
Cheng-Yu Hsieh, Chun-Liang Li, Chih-Kuan Yeh, Hootan Nakhost, Yasuhisa Fujii,
  Alexander Ratner, Ranjay Krishna, Chen-Yu Lee, and Tomas Pfister,
\newblock ``Distilling step-by-step! outperforming larger language models with
  less training data and smaller model sizes,''
\newblock {\em arXiv preprint arXiv:2305.02301}, 2023.

\bibitem{dinov22024}
Maxime Oquab, Timoth{\'e}e Darcet, Th{\'e}o Moutakanni, Huy Vo, Marc
  Szafraniec, Vasil Khalidov, Pierre Fernandez, Daniel Haziza, Francisco Massa,
  Alaaeldin El-Nouby, et~al.,
\newblock ``Dinov2: Learning robust visual features without supervision,''
\newblock {\em Transactions on Machine Learning Research Journal}, pp. 1--31,
  2024.

\bibitem{clipkd2024}
Chuanguang Yang, Zhulin An, Libo Huang, Junyu Bi, Xinqiang Yu, Han Yang, Boyu
  Diao, and Yongjun Xu,
\newblock ``Clip-kd: An empirical study of clip model distillation,''
\newblock in {\em Proceedings of the IEEE/CVF Conference on Computer Vision and
  Pattern Recognition}, 2024.

\bibitem{laion2022}
Christoph Schuhmann, Romain Beaumont, Richard Vencu, Cade Gordon, Ross
  Wightman, Mehdi Cherti, Theo Coombes, Aarush Katta, Clayton Mullis, Mitchell
  Wortsman, et~al.,
\newblock ``Laion-5b: An open large-scale dataset for training next generation
  image-text models,''
\newblock {\em Advances in Neural Information Processing Systems}, vol. 35, pp.
  25278--25294, 2022.

\bibitem{imagenet1k}
Olga Russakovsky, Jia Deng, Hao Su, Jonathan Krause, Sanjeev Satheesh, Sean Ma,
  Zhiheng Huang, Andrej Karpathy, Aditya Khosla, Michael Bernstein,
  Alexander~C. Berg, and Li~Fei-Fei,
\newblock ``{ImageNet Large Scale Visual Recognition Challenge},''
\newblock {\em International Journal of Computer Vision (IJCV)}, vol. 115, no.
  3, pp. 211--252, 2015.

\bibitem{gpt2020}
Tom Brown, Benjamin Mann, Nick Ryder, Melanie Subbiah, Jared~D Kaplan, Prafulla
  Dhariwal, Arvind Neelakantan, Pranav Shyam, Girish Sastry, Amanda Askell,
  et~al.,
\newblock ``Language models are few-shot learners,''
\newblock {\em Advances in neural information processing systems}, vol. 33, pp.
  1877--1901, 2020.

\bibitem{llava1.5}
Haotian Liu, Chunyuan Li, Yuheng Li, and Yong~Jae Lee,
\newblock ``Improved baselines with visual instruction tuning,''
\newblock in {\em Proceedings of the IEEE/CVF Conference on Computer Vision and
  Pattern Recognition}, 2024, pp. 26296--26306.

\bibitem{llama32024}
Abhimanyu Dubey, Abhinav Jauhri, Abhinav Pandey, Abhishek Kadian, Ahmad
  Al-Dahle, Aiesha Letman, Akhil Mathur, Alan Schelten, Amy Yang, Angela Fan,
  et~al.,
\newblock ``The llama 3 herd of models,''
\newblock {\em arXiv preprint arXiv:2407.21783}, 2024.

\bibitem{visualgenome}
Ranjay Krishna, Yuke Zhu, Oliver Groth, Justin Johnson, Kenji Hata, Joshua
  Kravitz, Stephanie Chen, Yannis Kalantidis, Li-Jia Li, David~A. Shamma,
  Michael~S. Bernstein, and Li~Fei-Fei,
\newblock ``Visual genome: Connecting language and vision using crowdsourced
  dense image annotations,''
\newblock {\em International Journal of Computer Vision}, vol. 123, pp. 32--73,
  2017.

\end{thebibliography}

\clearpage

\section{Supplementary Material}

\subsection{Implementation Details}
We introduced the main implementation details of our model in the main paper, here we provide a more comprehensive explanation.
\subsubsection{Optimization-based Textual Inversion (OTI)}
We conducted 350 iterations on the test split of the ImageNet dataset and validated on the validation split of the COCO dataset. Learning rates were swept over $\left \{2e-3, 2e-2, 5e-2\right \}$, and loss weights $\lambda_{OTIgpt}$ were explored in $\left \{0.5, 0.75, 1\right \}$, with models selected based on retrieval performance on the COCO validation set. An exponential moving average with a decay rate of 0.99 was applied. For an unlabeled image $I$, we utilized the zero-shot classification capability of CLIP to categorize it among the ~20K classes of the Open Images V7 dataset. A hyperparameter $k$was introduced to denote the $k$ distinct visual objects most similar to $I$, and we set $k=15$ for our experiments. Using a single A6000 GPU, OTI for CLIP (ViT-B/32 backbone) required approximately 108 seconds per image with a batch size of 32, and the entire iterative process took 101 hours to complete on a single A6000 GPU.

\subsubsection{Textual Inversion Network $\Theta$}
We train YUKINO for 50 epochs, sweeping learning rates over $\left \{1e-6, 1e-5, 1e-4\right \}$ and selecting models based on retrieval performance on the COCO validation set. The batch size is set to 256, and loss weights $\lambda_{gpt}$ are swept over $\left \{0.5, 0.75, 1\right \}$ to determine the best model. The temperature $\tau$ in Eq.~\ref{eq:loss} is fixed at 0.25. For each image, we set the number of associated visual objects $k$ to 150. Using a single A6000 GPU, YUKINO (with a ViT-B/32 backbone) processes approximately 2 seconds per image with a batch size of 256. Training the textual inversion network $\Theta$ takes 18 hours in total on a single A6000 GPU. Details of the textual inversion network $\Theta$ architecture are provided in Table~\ref{tab:thetaarchitecture}. For the B/32 and B/16 backbones, the dimensions of the CLIP feature space and token embedding space ($d$ and $d_w$) are both 512. For the L/14 backbone, $d$ and $d_w$ are both 768.

\subsubsection{Both OTI and $\Theta$}
We used the GPT-Neo-2.7B model, which has 2.7 billion parameters and was developed by EleutherAI, to generate the phrases for regularization. For each of the 20,932 class names in the Open Images V7 dataset , we pre-generated 256 phrases, with a temperature of 0.5 and a length constraint of a maximum of 35 tokens. The AdamW optimizer  was used with a weight decay of 0.01. Mixed precision was used to save memory and enhance computational efficiency.

\subsection{Evaluation Detail}
In both evaluation, we are using accuracy as our metric. However, due to the different composition of the datasets, there are differences in the way accuracy is calculated. Specifically, in SugarCREPE, the acc for an input is computed according to:
\begin{equation}
	acc(I, T^{t}, T^{n})=\left\{\begin{matrix}
		1 & s(T^{t},I) > s(T^{n}, I)\\
		0 & otherwise
	\end{matrix}\right.
\end{equation}

However, at Winoground, matching success requires a stronger compositional understanding of models. Concretely, given images $I^{t}$ and $I^{n}$ and captions $T^{t}$ and $T^{n}$, the text score for an example$(I^{t}, I^{n}, T^{t}, T^{n})$ is computed according to:
\begin{equation}
	f(I^{t},I^{n},T^{t},T^{n})=\left\{\begin{matrix}
		1 & s(T^{t},I^{t}) > s(T^{n},I^{t}) \\
		& \ and \ s(T^{n},I^{n}) > s(T^{t},I^{n})\\
		0 & otherwise
	\end{matrix}\right.
\end{equation}
The image score for an example is computed according to:
\begin{equation}
	g(I^{t},I^{n},T^{t},T^{n})=\left\{\begin{matrix}
		1 & s(T^{t},I^{t}) > s(T^{t},I^{n}) \\
		& \ and \ s(T^{n},I^{n}) > s(T^{n},I^{t})\\
		0 & otherwise
	\end{matrix}\right.
\end{equation}
The group score is:
\begin{equation}
	h(I^{t},I^{n},T^{t},T^{n})=\left\{\begin{matrix}
		1 &  f(I^{t},I^{n},T^{t},T^{n}) \\
		& \ and \ g(I^{t},I^{n},T^{t},T^{n}) \\
		0 & otherwise
	\end{matrix}\right.
\end{equation}
The $s(\cdot)$ is the similarity of the image/caption pair.

\begin{table}
	\centering
	\caption{Pytorch-style description of the textual inversion network $\Theta$. $d$ and $d_w$ represent the dimension of the CLIP feature space and token embedding space $\Gamma$, respectively.}
	\begin{tabular}{cc}
		Layer & Module \\ \midrule
		Input & nn.Linear($d$, $d*4$) \\
		GELU & nn.GELU \\
		Dropout & nn.Dropout(0.5) \\
		Hidden & nn.Linear($d*4$, $d*4$) \\
		GELU & nn.GELU \\
		Dropout & nn.Dropout(0.5) \\
		Output & nn.Linear($d*4$, $d_w$) \\ \bottomrule
	\end{tabular}
	\label{tab:thetaarchitecture}
\end{table}

\begin{table*}[!htbp]
	\centering
	\small
	\caption{Composition evaluations of the methods on various backbone. The Benchmark is SugarCREPE. Best and second-best scores are highlighted in bold and underlined, respectively.}
	\label{tab:i2t}
	\begin{tabular}{lc|ccccccc|ccc}
		\toprule
		\multirow{2}{*}{Dataset} & \multirow{2}{*}{Method} & \multicolumn{3}{c}{REPLACE} & \multicolumn{2}{c}{SWAP} & \multicolumn{2}{c|}{ADD} & \multicolumn{3}{c}{Avg}\\
		\cline{3-12}
		& & obj & att & rel & obj & att & obj & att & REPLACE & SWAP & ADD \\
		\midrule
		\multicolumn{12}{c}{Image Encoder: ViT-B-16} \\
		\midrule
		\multirow{5}{*}{DataComp}
		& CLIP & 92.68 & 79.82 & 63.94 & 56.33 & 57.66 & 84.34 & 78.61 & 78.81 & 57.00 & 81.48 \\
		& Neg-CLIP & 89.10 & 82.99 & 69.35 & 68.98 & 69.82 & 82.78 & 81.50 & 80.48 & 69.40 & 82.14 \\
		& Structure-CLIP & 90.68 & 85.66 & 74.89 & 76.33 & 78.38 & 85.84 & 88.29 & 83.74 & 77.36 & 87.07 \\
		& YUKINO-OTI (Ours) & \underline{96.00} & \underline{93.15} & \underline{88.34} & \underline{84.49} & \underline{84.53} & \underline{93.89} & \underline{92.77} & \underline{92.50} & \underline{84.51} & \underline{93.33} \\
		& YUKINO (Ours) & \textbf{97.58} & \textbf{95.94} & \textbf{93.10} & \textbf{89.39} & \textbf{90.09} & \textbf{96.27} & \textbf{96.24} & \textbf{95.54}  & \textbf{89.74}  & \textbf{96.26}  \\
		\midrule
		\multicolumn{12}{c}{Image Encoder: ViT-B-32} \\
		\midrule
		\multirow{5}{*}{LAION2B}
		& CLIP & 93.77 & 82.49 & 68.92 &  60.00 & 67.42 & 87.10 & 77.89 & 81.73 & 63.71 & 82.50 \\
		& Neg-CLIP & 92.62 & 84.64 & 71.41 & 74.69 & 76.58 & 86.71 & 85.26 & 82.89 & 75.64 & 85.99 \\
		& Structure-CLIP & 91.77 & 86.8 & 75.32 & 74.29 & 84.83 & 88.94 & 88.01 & 84.63 & 79.56 & 88.48 \\
		& YUKINO-OTI (Ours) & \textbf{97.82} & \textbf{95.43} & \textbf{91.68} & \underline{87.76} & \textbf{90.69} & \underline{96.51} & \textbf{94.80} & \textbf{94.98} & \textbf{89.23} & \textbf{95.66} \\
		& YUKINO (Ours) & \underline{97.15} & \underline{94.80} & \underline{91.25} & \textbf{88.57} & \underline{89.79} & \textbf{96.65} & \underline{93.21} & \underline{94.40}  & \underline{89.18}  & \underline{94.93}  \\
		\midrule
		\multicolumn{12}{c}{Image Encoder: ViT-L-14} \\
		\midrule
		\multirow{5}{*}{WIT}
		& CLIP & 94.07 & 79.19 & 65.15 & 60.41 & 62.31 & 78.32 & 71.53 & 79.47 & 61.36 & 74.925 \\
		& Neg-CLIP & 92.86 & 81.98 & 75.04 & 75.51 & 74.17 & 88.31 & 85.40 & 83.29 & 74.84 & 86.86 \\
		& Structure-CLIP & 95.04 & 87.56 & 77.38 & 76.33 & 81.98 & 90.35 & 88.73 & 86.66 & 79.16 & 89.54 \\
		& YUKINO-OTI (Ours) & \underline{98.37} & \underline{96.45} & \underline{94.45} & \underline{91.43} & \underline{91.14} & \underline{95.64} & \underline{93.79} & \underline{96.42} & \underline{91.29} & \underline{94.72} \\
		& YUKINO (Ours) & \textbf{99.09} & \textbf{97.84} & \textbf{95.66} & \textbf{95.51} & \textbf{95.95} & \textbf{96.80} & \textbf{97.11} & \textbf{97.53}  & \textbf{95.73}  & \textbf{96.96}  \\
		\bottomrule
	\end{tabular}
\end{table*}

\subsection{Detailed Results}
Previously, we reported results on the SugarCREPE and Winoground benchmarks. However, to demonstrate the effectiveness of our approach across different backbones, we also evaluated other backbones, such as ViT-B/16 and ViT-L/14.

In Table~\ref{tab:i2t}, our approach demonstrates optimal performance across all backbones. DataComp has superior data quality compared to LAION2B, and ViT-L/14 has a more complex architecture than ViT-B/32, which enables the visual encoder to capture higher-level visual semantics more effectively. Additionally, our YUKINO model, trained on pre-generated pseudo-tokens, further enhances this ability, resulting in stronger compositionality.

In Table~\ref{tab:t2i}, we present the results of our method on Winoground. Our approach achieves the best performance in terms of Text Score, Image Score and Group Score. We added two new metrics to Winoground: Single Image Score and Single Text Score. The Single Image Score evaluates one image with two text captions, simplifying Winoground's metric to a level comparable with SugarCREPE. Compared to CLIP, Neg-CLIP and Structure-CLIP show a slight drop in Text Score and Single Image Score. However, YUKINO delivers improvements across all backbones. As described by SugarCREPE\cite{sugarcrepe2024}, the unintentionally overfitting of the model overestimates the improvements in compositional understanding.

Moreover, the Single Text Score to Image Score of Structure-CLIP under the ViT-B-32 backbone decreased by 41.00$\%$, and Neg-CLIP decreased by 42.25$\%$. However, YUKINO showed a smaller decrease of only 25.50$\%$. Models fine-tuned on hard negative samples show improved scores when performing tasks like multiple-choice questions (similar to the evaluation in SugarCREPE). However, under a mixed modality, the performance of these supervised models significantly drops. This is consistent with the conclusions drawn from our representation analysis.

\begin{table*}[htbp]
	\centering
	\small
	\caption{Composition evaluations of the methods on various backbone. The Benchmark is Winoground. Best and second-best scores are highlighted in bold and underlined, respectively.}
	\label{tab:t2i}
	\begin{tabular}{lc|ccc|cc}
		\toprule
		Dataset & Method & Text Score & Image Score & Group Score & Single Image Score & Single Text Score \\
		\midrule
		\multicolumn{7}{c}{Image Encoder: ViT-B-16} \\
		\midrule
		\multirow{5}{*}{DataComp}
		& CLIP & 26.25 & 8.00 & 6.00 & 56.75 & 53.13 \\
		& Neg-CLIP & 20.75 & 10.00 & 6.00 & 55.00 & 53.75 \\
		& Structure-CLIP & 19.25 & 10.50 & 6.00 & 55.63 & 53.88 \\
		& YUKINO-OTI (Ours) & \underline{51.25} & \underline{26.00} & \underline{22.00} & \underline{68.50} & \underline{60.38} \\
		& YUKINO (Ours) & \textbf{66.25} & \textbf{36.50} & \textbf{31.50} & \textbf{79.63} & \textbf{66.88} \\
		\midrule
		\multicolumn{7}{c}{Image Encoder: ViT-B-32} \\
		\midrule
		\multirow{5}{*}{LAION2B}
		& CLIP & 34.75 & 11.00 & 7.50 & 62.38 & 54.13 \\
		& Neg-CLIP & 20.50 & 12.25 & 6.50 & 57.13 & 54.50 \\
		& Structure-CLIP & 22.50 & 14.00 & 8.25 & 57.63 & 55.00\\
		& YUKINO-OTI (Ours) & \underline{66.75} & \underline{42.50} & \underline{37.50} & \underline{81.00} & \underline{69.75} \\
		& YUKINO (Ours) & \textbf{74.00} & \textbf{46.25} & \textbf{42.75} & \textbf{84.50} & \textbf{71.75} \\
		\midrule
		\multicolumn{6}{c}{Image Encoder: ViT-L-14} \\
		\midrule
		\multirow{5}{*}{WIT}
		& CLIP & 28.75 & 11.00 & 8.50 & \underline{60.13} & 55.00 \\
		& Neg-CLIP & 22.50 & 12.00 & 8.00 & 58.00 & 54.63 \\
		& Structure-CLIP & 25.25 & 16.25 & 10.50 & 57.25 & \underline{55.38} \\
		& YUKINO-OTI (Ours) & \underline{34.50} & \underline{17.50} & \underline{13.50} & 54.75 & 52.50 \\
		& YUKINO (Ours) & \textbf{45.50} & \textbf{29.50} & \textbf{22.50} & \textbf{64.13} & \textbf{60.88} \\
		\bottomrule
	\end{tabular}
\end{table*}

\subsection{Case Study}
\begin{figure*}[!htbp]
	\centering
	\includegraphics[width=0.95\linewidth]{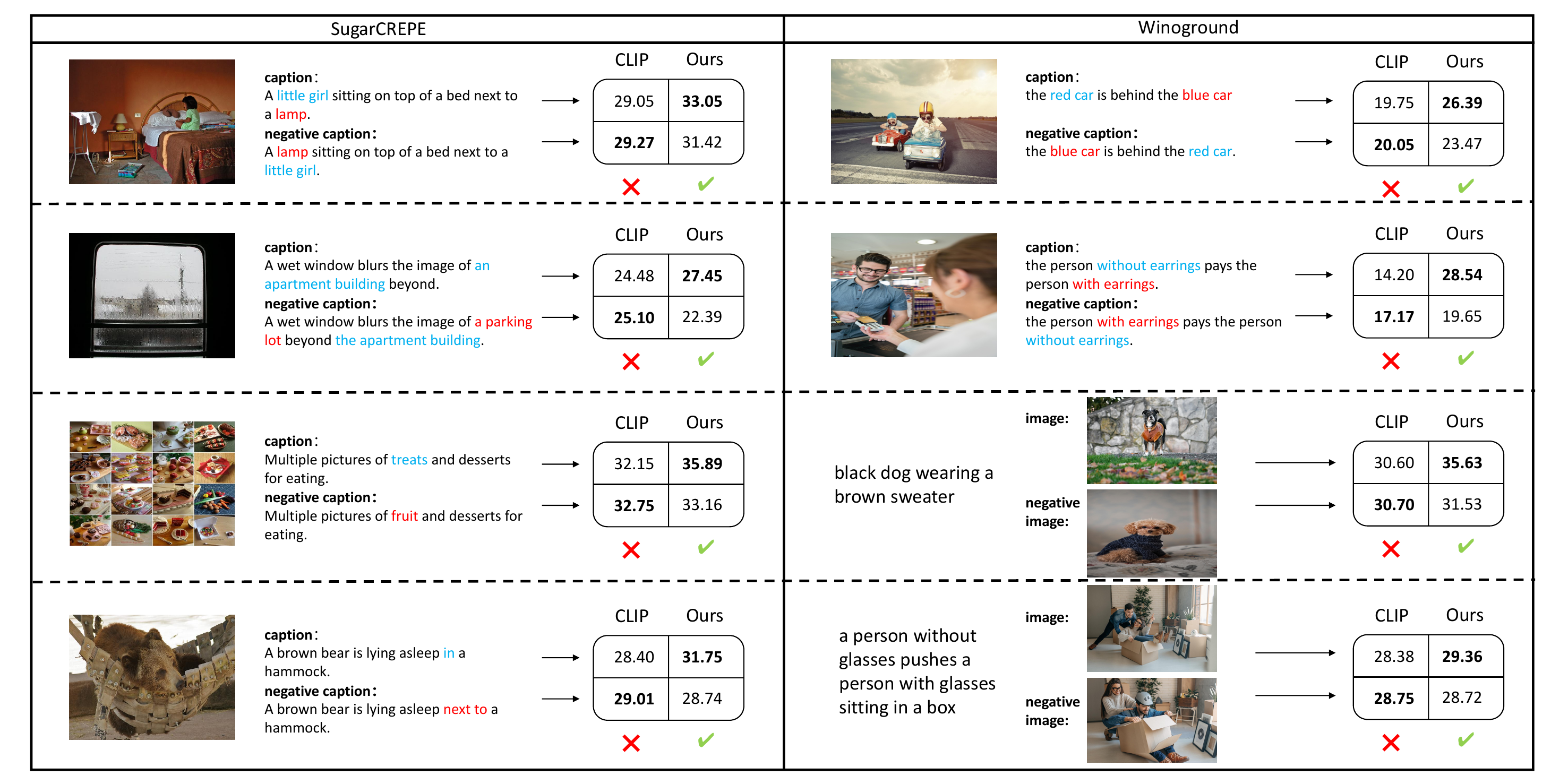}
	\caption{Predictions of different approaches. The words in red and blue are difference words. We compare our YUKINO with CLIP to calculate CLIP scores (i.e., semantic similarity) between the image and captions.
	}
	\label{fig:case}
	\vspace{-10pt}
\end{figure*}

To validate the performance of the proposed model, visual case studies were conducted by selecting eight pairs of samples from SugarCREPE and Winoground. The prediction results of the cases are presented in Figure ~\ref{fig:case}, which illustrates that YUKINO can successfully distinguish between positive and negative samples. However, the CLIP model encounters challenges in accurately determining the semantic similarities between these captions and the given image. In particular, under the Text-to-Image formulation,  the CLIP model demonstrates nearly uniform semantic similarity, indicating a lack of compositional understanding. In contrast to the CLIP model, YUKINO exhibits sensitivity to modifications in fine-grained semantics, indicating its ability for compositional understanding. As an example, the caption ``A little girl sitting on top of a bed next to a lamp'' is used to evaluate the ability of YUKINO to distinguish between positive and negative samples when two objects (i.e., little girl and lamp) are exchanged. The results show that YUKINO can make a distinction between positive and negative captions with a margin of 1.63$\%$, which further verifies the effectiveness of the proposed method in enhancing multi-modal structured representations.

\end{document}